\documentclass[letterpaper, 10 pt, journal, twoside]{IEEEtran}

\IEEEoverridecommandlockouts                              % This command is only needed if 
\usepackage{booktabs}
\usepackage{amsmath} % assumes amsmath package installed
\usepackage{amssymb}  % assumes amsmath package installed
\usepackage{cite}
\usepackage[vlined,linesnumbered,ruled,noend]{algorithm2e}
\usepackage{graphicx}
\usepackage{todonotes}
\usepackage{multirow}
\newcommand{\etal}{\emph{et al.}}
\usepackage[bookmarks=true]{hyperref}

\usepackage[top=58pt, bottom=60pt, left=53pt, right=53pt]{geometry}
\usepackage[skip=0pt,labelfont=bf]{caption}

\usepackage{subfig}
\usepackage{etoolbox}
\makeatletter
\patchcmd{\@makecaption}
{\scshape}
{}
{}
{}
\makeatletter
\patchcmd{\@makecaption}
{\\}
{.\ }
{}
{}
\makeatother
\def\tablename{Table}

\newcommand{\shrinka}{\def\baselinestretch{0.98}\large\normalsize}

\title{Benchmarking In-Hand Manipulation}
\author{Silvia Cruciani*$^1$, Balakumar Sundaralingam*$^2$, Kaiyu Hang$^3$,  Vikash Kumar$^4$, \\Tucker Hermans$^{2,5}$, and Danica Kragic$^1$
  \thanks{Manuscript received: August, 15, 2019; Revised November, 18, 2019; Accepted December, 1, 2019.}
  \thanks{This paper was recommended for publication by Editor Han Ding upon evaluation of the Associate Editor and Reviewers' comments.}
  \thanks{S.~Cruciani was supported by Swedish  Foundation  for Strategic  Research  project  GMT14-0082  FACT and B.~Sundaralingam was supported by NSF Award \#1846341}   
    \thanks{\small *These two authors contributed equally.}
  \thanks{$^{1}$Division of Robotics, Perception \& Learning, EECS, KTH Royal Institute of Technology, Stockholm, Sweden}
  \thanks{$^{2}$Robotics Center \& School of Computing, University of Utah, UT USA.}
  \thanks{$^{3}$Dept. of Mechanical Engineering \& Material Science, Yale University, New Haven, CT, USA.}
  \thanks{$^{4}$Google AI}
  \thanks{$^{5}$NVIDIA Research}
}
\begin{document}
\shrinka
\maketitle
%\shrinka
%\thispagestyle{empty}
%\pagestyle{empty}
\begin{abstract}
The purpose of this benchmark is to evaluate the planning and control aspects
of robotic in-hand manipulation systems. The goal is to assess the system's ability to change the pose of a hand-held object by either using the fingers, environment or a combination of both. Given an object surface mesh from the YCB data-set, we provide examples of initial and goal states (i.e.\ static object poses and fingertip locations) for various in-hand manipulation tasks.
We further propose metrics that measure the error in reaching the goal state from a specific initial state, which, when aggregated across all tasks, also serves as a measure of the system's in-hand manipulation capability. We provide supporting software, task examples, and evaluation results associated with the benchmark.
\end{abstract}

\section{Introduction}
\begin{figure}[t]
    \centering
    \subfloat{ %
        \includegraphics[height=4.0cm]{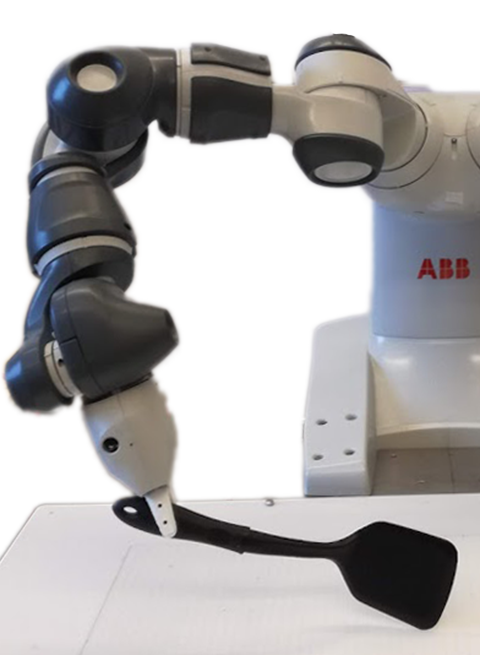}%
    }%
    \subfloat{ %
        \includegraphics[height=4cm]{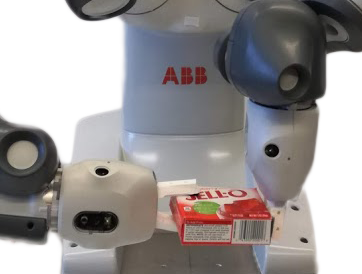}%
      }
      \\ \vspace{0.3cm}
    \includegraphics[width=0.48\textwidth]{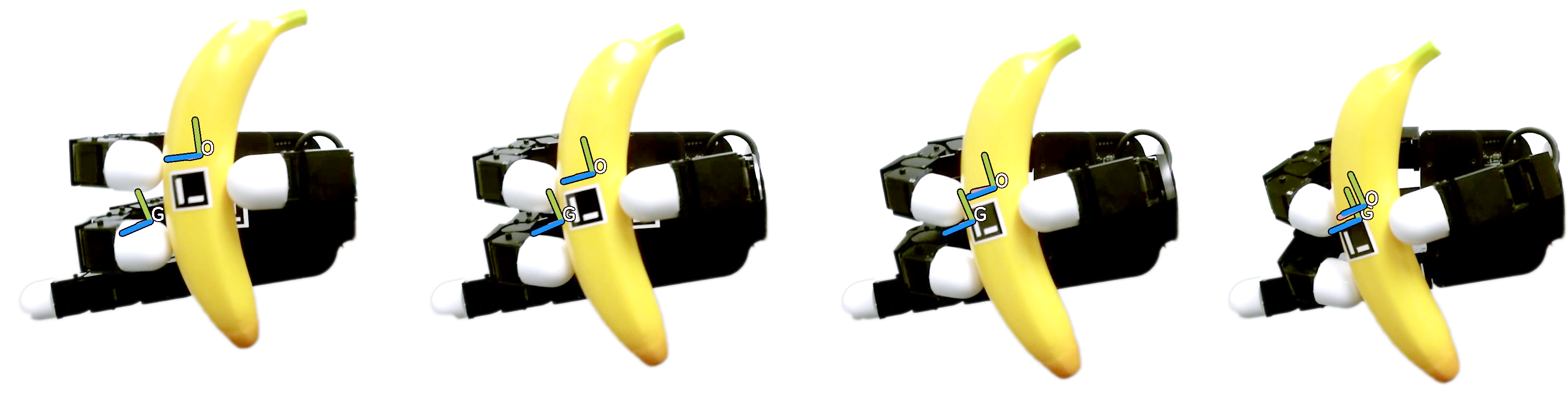}
    
    \caption{In-hand manipulation strategy examples: \emph{top left:} pushing against the environment; \emph{top right:} bi-manual push; \emph{bottom:} exploitation of dexterity in a multi-finger hand.}
    \label{fig:in-hand_solutions_example}
\end{figure}

\IEEEPARstart{I}{n-hand} manipulation is the task of changing the grasp on an hand-held object without placing it back and picking it up again. In recent years, researchers have demonstrated in-hand manipulation skills on real robotic platforms with end-effectors ranging from simple grippers~\cite{chavan-dafle_extrinsic_dexterity} to anthropomorphic hands~\cite{kumar2016optimal,van2015learning,andrychowicz_learning_dexterous_manipulation,higo2018rubik} on everyday objects~(some examples shown in Fig.~\ref{fig:in-hand_solutions_example}). This growing interest in in-hand manipulation has created the need for a thorough comparison of the proposed methods using a common set of tasks and metrics. However, the diversity in the methodological approaches and hardware used makes the burden of comparison rather complex for individual research labs to perform. There is thus a need for a benchmarking protocol that allows researchers to evaluate new methodologies in a more standardized manner.

In this work, we propose a benchmarking scheme for in-hand manipulation that can extend to all kinds of robotic systems. All the supporting material is available at \url{https://robot-learning.cs.utah.edu/project/benchmarking_in_hand_manipulation}.

While some benchmark examples are available for evaluating grasp planning in a simulated environment~\cite{ulbrich_openGRASP_benchmark, kootstra_visgrab}, we believe that for contact-rich tasks such as grasping and in-hand manipulation it is of fundamental importance to evaluate the performance of the system in a physical, real-world scenario. Therefore, we propose a series of tasks to evaluate the capabilities of a robotic system to plan and execute in-hand manipulation using a set of objects available from the Yale-CMU-Berkeley (YCB) object and model set~\cite{calli_ycb}. This set provides both object meshes and physical objects, so that different research groups can rely on the same test set. 

While our benchmark focuses on physical robot execution, the procedure is still valid in simulation and also for benchmarking of planning algorithms. The availability of object meshes for the YCB data-set enables easy evaluation in common robotics simulators. As such, we will host results for simulation in a separate section of the associated website. 

%%% Local Variables:
%%% mode: latex
%%% TeX-master: "main"
%%% End:

To motivate our choice of protocol and metrics, we briefly review the relevant in-hand manipulation literature with a focus on how researchers perform evaluation and validation of the proposed methods.
Active research on in-hand manipulation explores a wide variety of robotic systems, ranging from grippers~\cite{chavan-dafle_in-hand_manipulation_motion_cones,cruciani_dexterous_manipulation_graph} to dexterous hands~\cite{sundaralingam-auro2018-in-grasp-optimization,kumar2016optimal}. Researchers have also augmented mobile robots~\cite{rus1997coordinated} and multiple arms to behave as fingertips~\cite{li2013integrating,han_dexterous_manipulation,lowrey2018reinforcement} to perform in-hand manipulation.  Methodologies for analyzing the dexterity of robotic hands were developed in~\cite{roa2014:evaluation_manipulation} and \cite{odner2011:dexterous_manipulation_underactuated}; however, they focus on the hands' kinematic reachability and actuation, rather than on execution with different objects and tasks.

% multi-fingered,underactuated manipulation
With multi-fingered dexterous hands, in-hand manipulation has been performed leveraging the redundancy in the fingers to move the object without completely releasing the grasp~\cite{michelman1993compliant,rus1999hand,sundaralingam-icra2018-finger-gaiting, sundaralingam-auro2018-in-grasp-optimization, psomopoulou_stable_pinching,andrychowicz_learning_dexterous_manipulation,kumar2016optimal}. For under-actuated hands, model based control has been successfully employed%to perform in-hand manipulation
~\cite{calli2017vision,liarokapis_dexterous_adaptive_hands, rojas_underactuated_hand_for_in-hand, chavan-dafle_shape-shifting_gripper}. Alternatively, task-specific design of under-actuated hands enables a limited set of re-positioning possibilities~\cite{rahman_dexterous_gripper, bircher_2fingered_gripper_for_reorientation,spiers2018variable}. Researchers have leveraged learning methods to manipulate objects~\cite{van2015learning,calli2018learning,antonova_rl_pivoting}, and validated machine learning approaches in physics simulation~\cite{Rajeswaran-RSS-18,akrour2018regularizing}. In-hand manipulation has also been explored as a planning problem~\cite{sundaralingam-icra2018-finger-gaiting,mordatch2012discovery}.

% grippers
With parallel grippers, in-hand manipulation has been achieved
by exploiting the concept of \emph{extrinsic dexterity}~\cite{chavan-dafle_extrinsic_dexterity}, in which the degrees of freedom of a certain gripper are enhanced by means of external supports such as contact with the environment, gravity and friction. Some of these works exploit gravity and inertial forces to initiate the motion of the object~\cite{vina_adaptive_control_pivoting,sintov_swing-up_regrasping, cruciani_3stages_pivoting,antonova_rl_pivoting,shi_dynamic_sliding_journal}, while others rely on pushes against an external contact. This contact can be either a fixture or a surface present in the environment~\cite{chavan-dafle_in-hand_manipulation_motion_cones, hou_fast_planning_for3D_any-pose-reorientation, almeida_dexterous_manipulation_external_contacts}, or a second gripper in a dual-arm manipulation scenario~\cite{cruciani_dexterous_manipulation_graph}.

% grasp stability, simulation
One of the main challenges in in-hand manipulation is balancing the object in a stable configuration during the execution. An alternative to keep stability is dynamic in-hand manipulation by tossing the object in the air and catching it in the desired pose~\cite{furukawa_dynamic_regrasping,chavan-dafle_extrinsic_dexterity}. A more conservative approach is to balance the object with extra support during execution. Researchers have used a planar surface~\cite{higo2018rubik,katzschmann2015autonomous} and even the palm of the hand~\cite{andrychowicz_learning_dexterous_manipulation} to balance the object.

Across these various methods for in-hand manipulation, we found that the goal was defined as either a target pose~\cite{sundaralingam-auro2018-in-grasp-optimization,andrychowicz_learning_dexterous_manipulation,chavan2018stable}, desired contact locations on the object~\cite{sundaralingam-icra2018-finger-gaiting}, or both~\cite{cruciani_dexterous_manipulation_graph}. The target pose can range from single dimensional rotation~\cite{cruciani_3stages_pivoting}, through planar~\cite{van2015learning,calli2018learning} to full 3D poses~\cite{sundaralingam-auro2018-in-grasp-optimization}. When desired contacts are defined, the goal is the position of each specific contact point on the object that the robot needs to reach~\cite{sundaralingam-icra2018-finger-gaiting}. A combination of target pose and desired contact is commonly user-specified~\cite{cruciani_dexterous_manipulation_graph}. We summarize the related work in terms of the goal definition, use of extra object support, and real-world execution in Table~\ref{tab:related_work}.

Recently, Rajeswaran~\etal~\cite{Rajeswaran-RSS-18} proposed a set of manipulation tasks to benchmark dexterous manipulation. The benchmark primarily focuses on simulated dexterous hands. The proposed tasks do not account for all kinds of manipulation platforms~(e.g. the proposed hammer task may be impractical with a parallel gripper). They also do not propose qualitative error metrics to measure the capabilities of a system. In contrast, we design platform-agnostic in-hand manipulation tasks and we propose qualitative metrics to evaluate the system's performance.

We organize the remainder of this letter as follows. We describe our protocol in~Sec.~\ref{sec:protocol_design}, followed by a discussion of the benchmarking guidelines in~Sec.~\ref{sec:guidelines}. We demonstrate example results for representative methods of in-hand manipulation, benchmarked using the proposed protocol in Sec.~\ref{sec:demo}, and conclude the letter in~Sec.~\ref{sec:conclusion}.

\begin{table}
  \centering
  \begin{tabular}{l c c r r}
    \toprule
    \multirow{2}{*}{\textbf{Article}}&\multicolumn{2}{c}{\textbf{Goal}} & \multirow{2}{*}{\textbf{Obj. support}} & \multirow{2}{*}{\textbf{Real-world}} \\ \cline{2-3}                              &
                                                                                                                                                                                                            \textbf{Pose} & \textbf{Contact}  & \\ \toprule

    \cite{vina_adaptive_control_pivoting,cruciani_3stages_pivoting,sintov_swing-up_regrasping} & 1D rot. & & No & Yes    \\ \midrule
    \cite{van2015learning,calli2018path,calli2018robust}  & 2D tran. &  & No & Yes\\ \midrule
    \cite{spiers2018variable,lowrey2018reinforcement} & SE(2) &   & Ext. surface & Yes\\ \midrule
    \cite{shi_dynamic_sliding_journal,chavan2018stable,chavan-dafle_in-hand_manipulation_motion_cones} & SE(2) & & No & Yes \\ \midrule
    \cite{higo2018rubik}&  \multicolumn{2}{c}{SE(3)\hspace{8pt} Designed}  & Ext. surface & Yes\\ \midrule
    \cite{kumar2016optimal,andrychowicz_learning_dexterous_manipulation} & SE(3) & &  Palm & Yes \\ \midrule
    \cite{sundaralingam_in-grasp_manipulation,sundaralingam-auro2018-in-grasp-optimization} & SE(3) &  & No & Yes \\ \midrule
    \cite{cruciani_dexterous_manipulation_graph} & SE(3) & Points & No & Yes \\ \midrule
    \cite{sundaralingam-icra2018-finger-gaiting} &  & Points &  NA & No\\ \midrule
    \cite{Rajeswaran-RSS-18} & \multicolumn{2}{c}{SE(3) Demonstration} & NA & No\\
    \bottomrule
  \end{tabular}
  \caption{Research articles categorized in terms of goal definition and other relevant features to protocol formulation.}
  \label{tab:related_work}
\end{table}
%%% Local Variables:
%%% mode: latex
%%% TeX-master: "main"
%%% End:

\section{Protocol Design}\label{sec:protocol_design}
Our objective is to assess the capability of a system to execute in-hand manipulation tasks in a goal-directed manner. We define an in-hand manipulation task as the problem of changing the grasp on an object without placing it on a support surface. In this given task, the goal is to change the object's configuration inside the robot's hand, from an initial grasp to a final desired grasp. A grasp, $G$, is defined by the hand pose,~$H$, with respect to the object's frame and the contact points,~$P$, made between the hand and the object. Since different robot geometries will require different hand poses and contact points to successfully grasp the object of interest, we specify the protocol in such a way that it can be adapted to different grasps according to the available setup.

We highlight that the specification of $H$ w.r.t. the object is equivalent to specifying the object's pose inside the hand. We do not constrain the \emph{absolute} object's pose in a different, fixed reference frame because the focus is on the \emph{relative} pose between the hand and the object.

\subsection{Task Description}
\begin{figure}
\centering
\includegraphics[trim={0 0.3cm 1cm 0}, clip,width=0.46\textwidth]{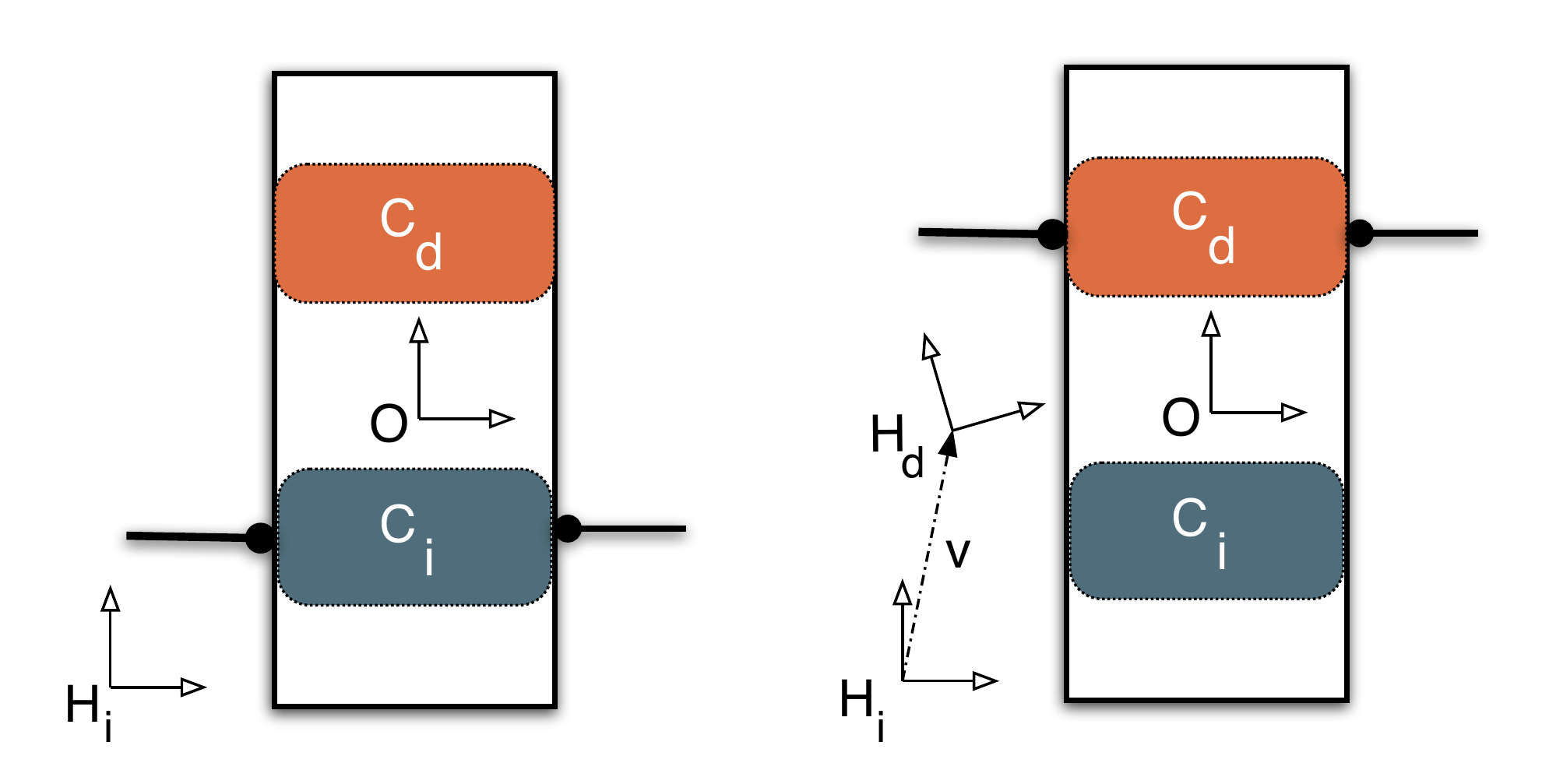}
\caption{We illustrate the robot's fingertips as lines with solid circles defining contact with the rectangular object. We illustrate the task in 2D for clarity, but define it in 3D. Frames~$H_i$ and~$H_d$ define the hand's initial and desired final poses respectively. The left image shows the initial task setup, while the right shows a solution reaching the desired state.}
\label{fig:task_illustration}
\end{figure}

\begin{figure}
\centering
 \subfloat{ %
        \includegraphics[trim={400 150 200 100},clip,width=0.12\textwidth]{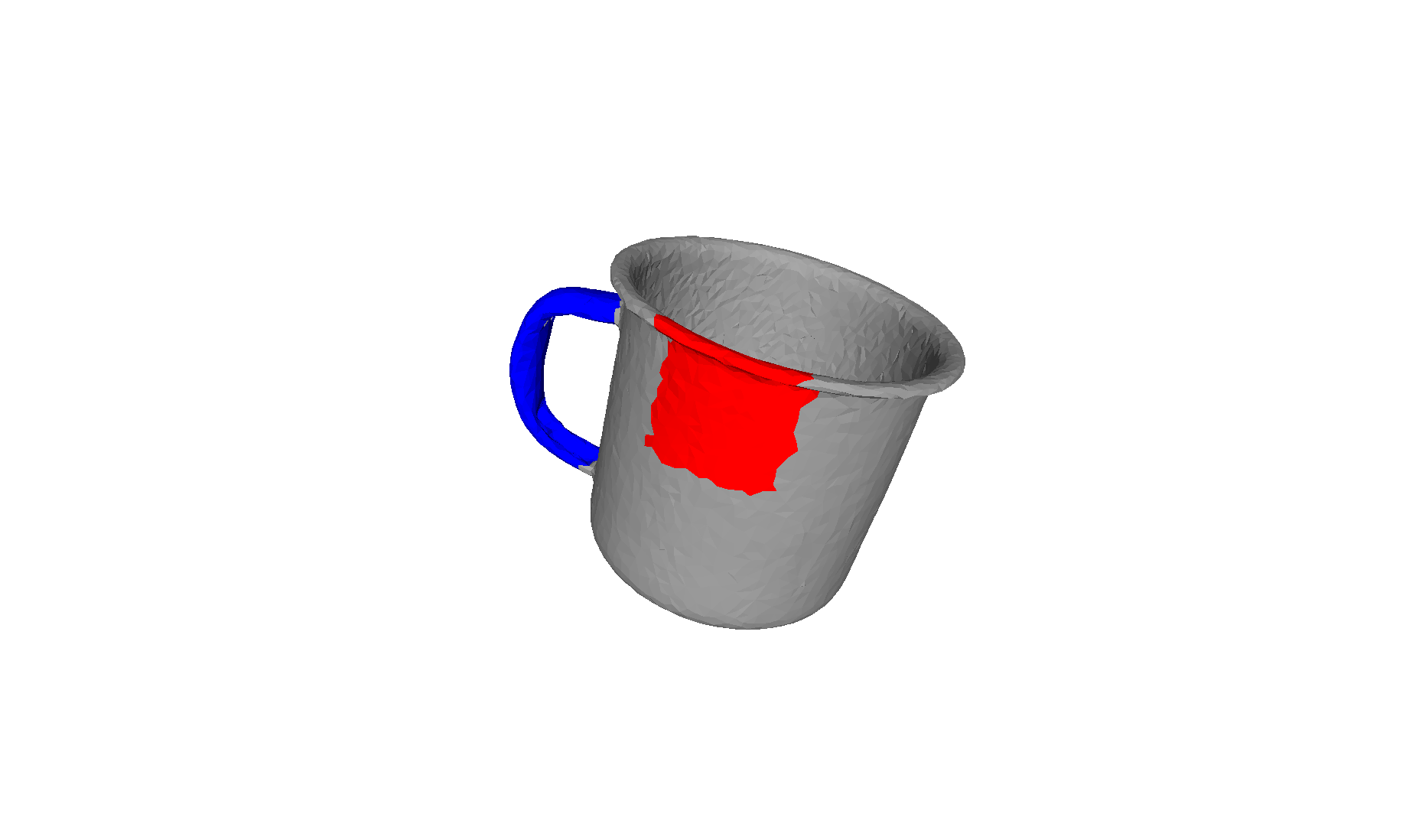}%
    }%
    \subfloat{ %
        \includegraphics[trim={200 100 300 100},clip,width=0.12\textwidth]{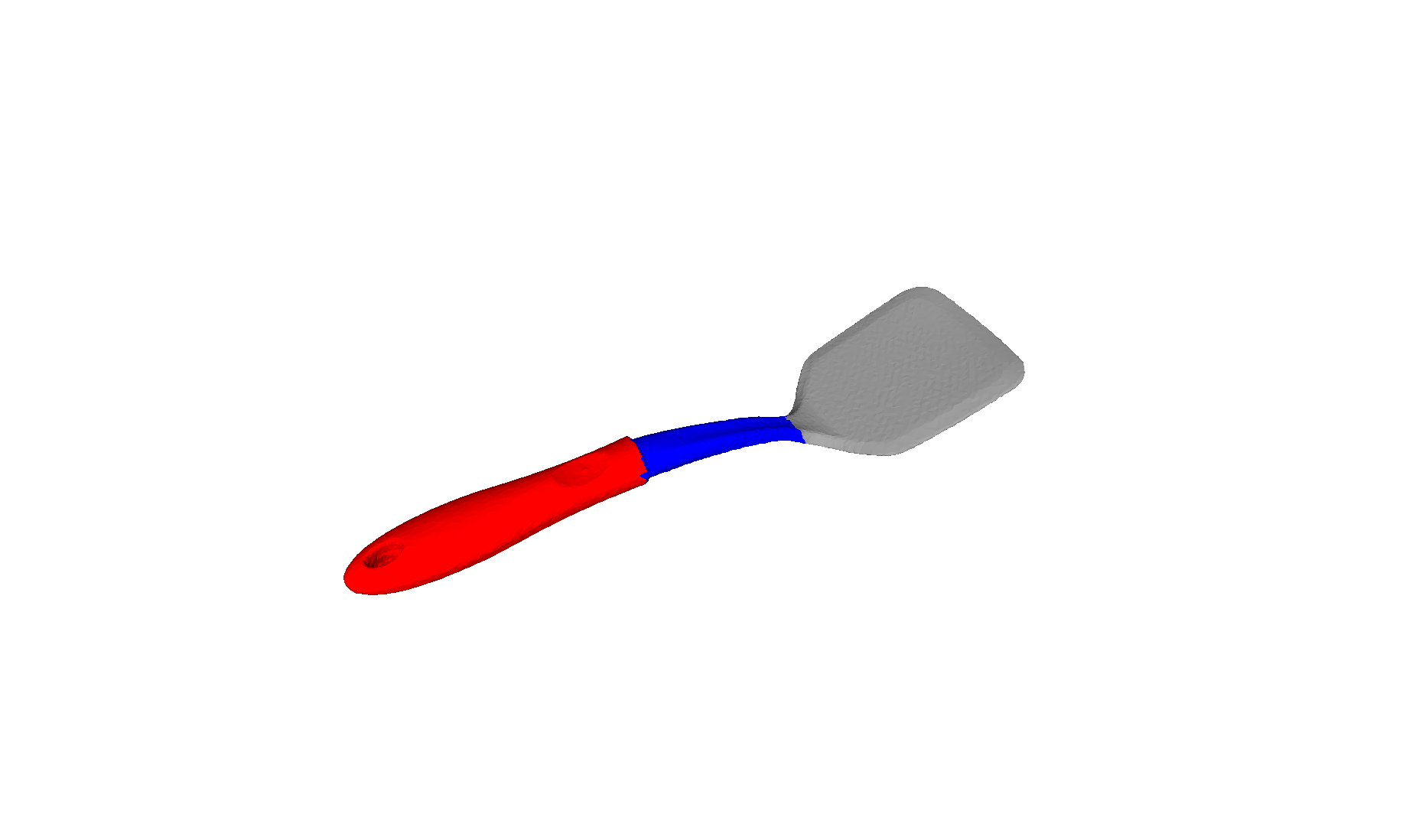}%
    }
    \subfloat{ %
        \includegraphics[trim={400 100 200 100},clip,width=0.12\textwidth]{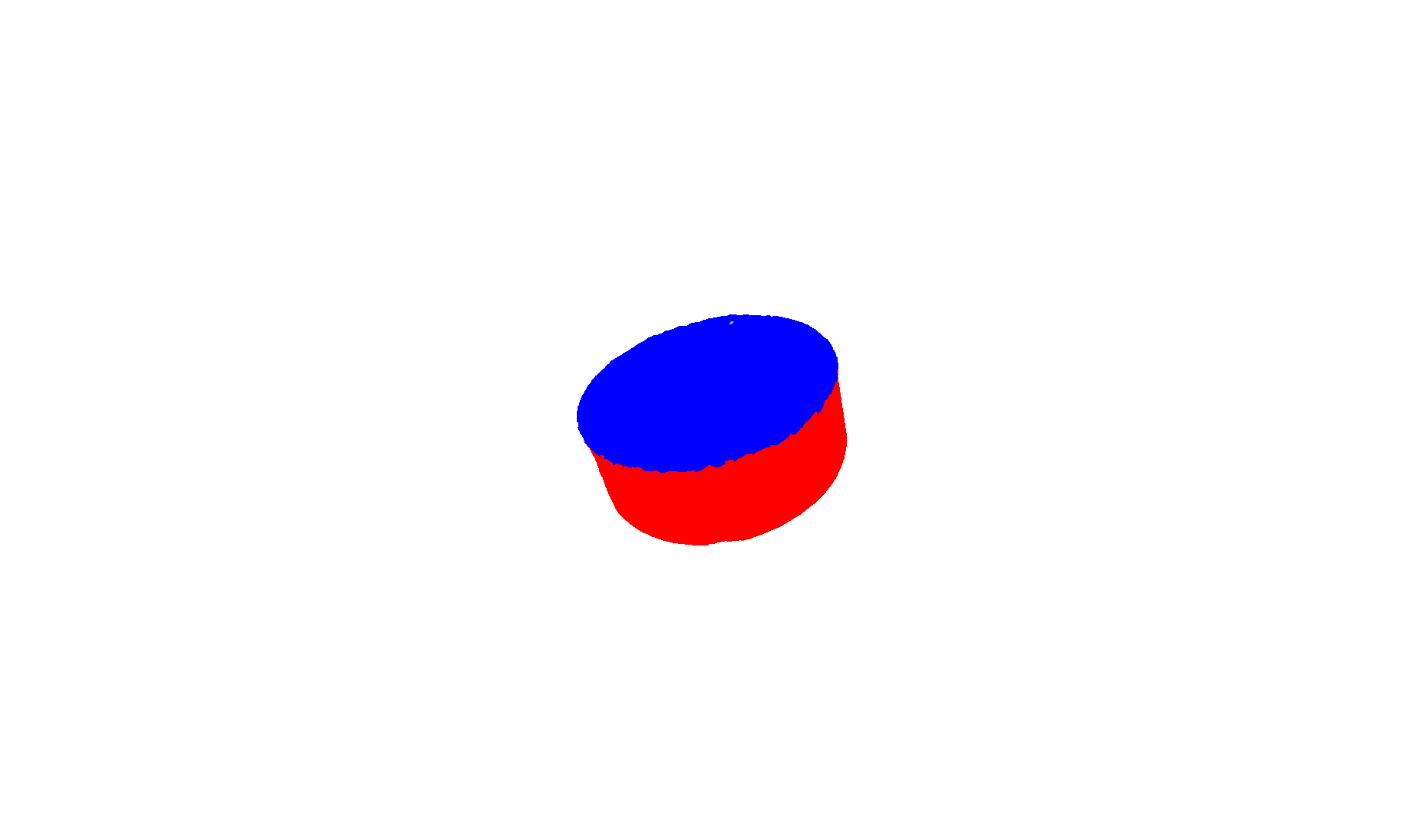}%
    }
    \subfloat{ %
        \includegraphics[trim={400 100 200 100},clip,width=0.12\textwidth]{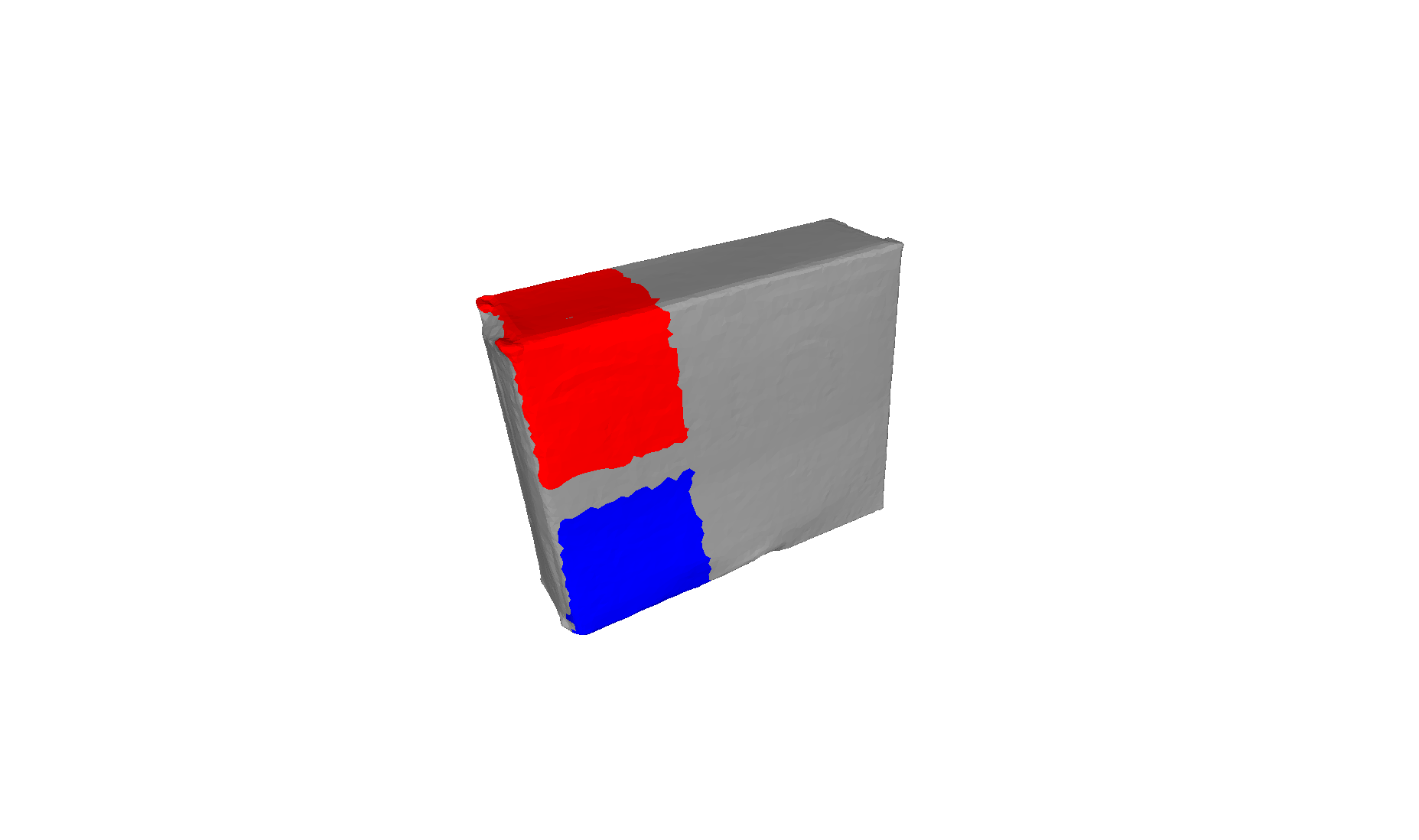}%
    }\\[-0.5cm]
    \subfloat{ %
        \includegraphics[trim={400 100 200 100},clip,width=0.12\textwidth]{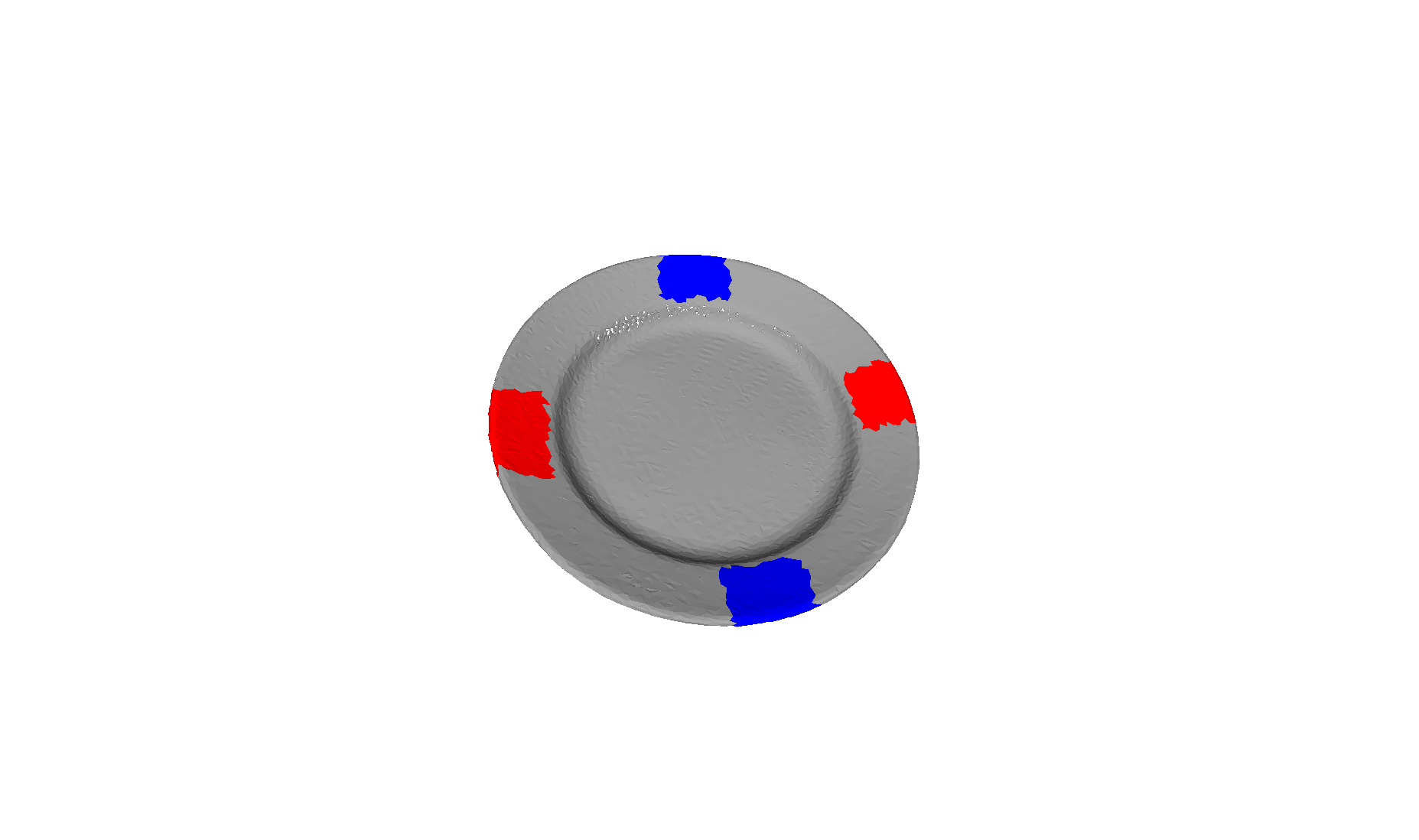}%
    }
    \subfloat{ %
        \includegraphics[trim={400 100 200 100},clip,width=0.12\textwidth]{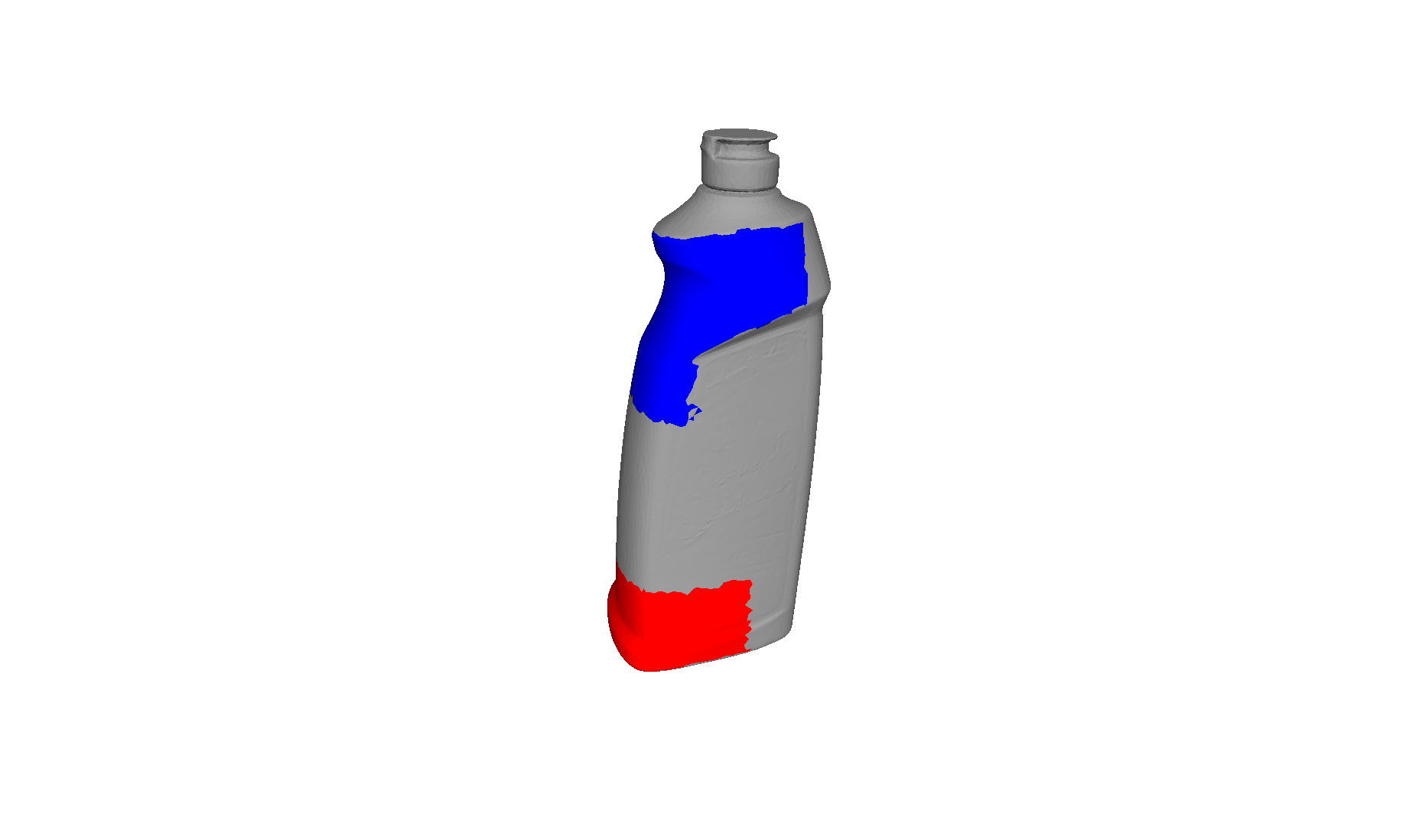}%
    }
    \subfloat{ %
        \includegraphics[trim={400 100 200 100},clip,width=0.12\textwidth]{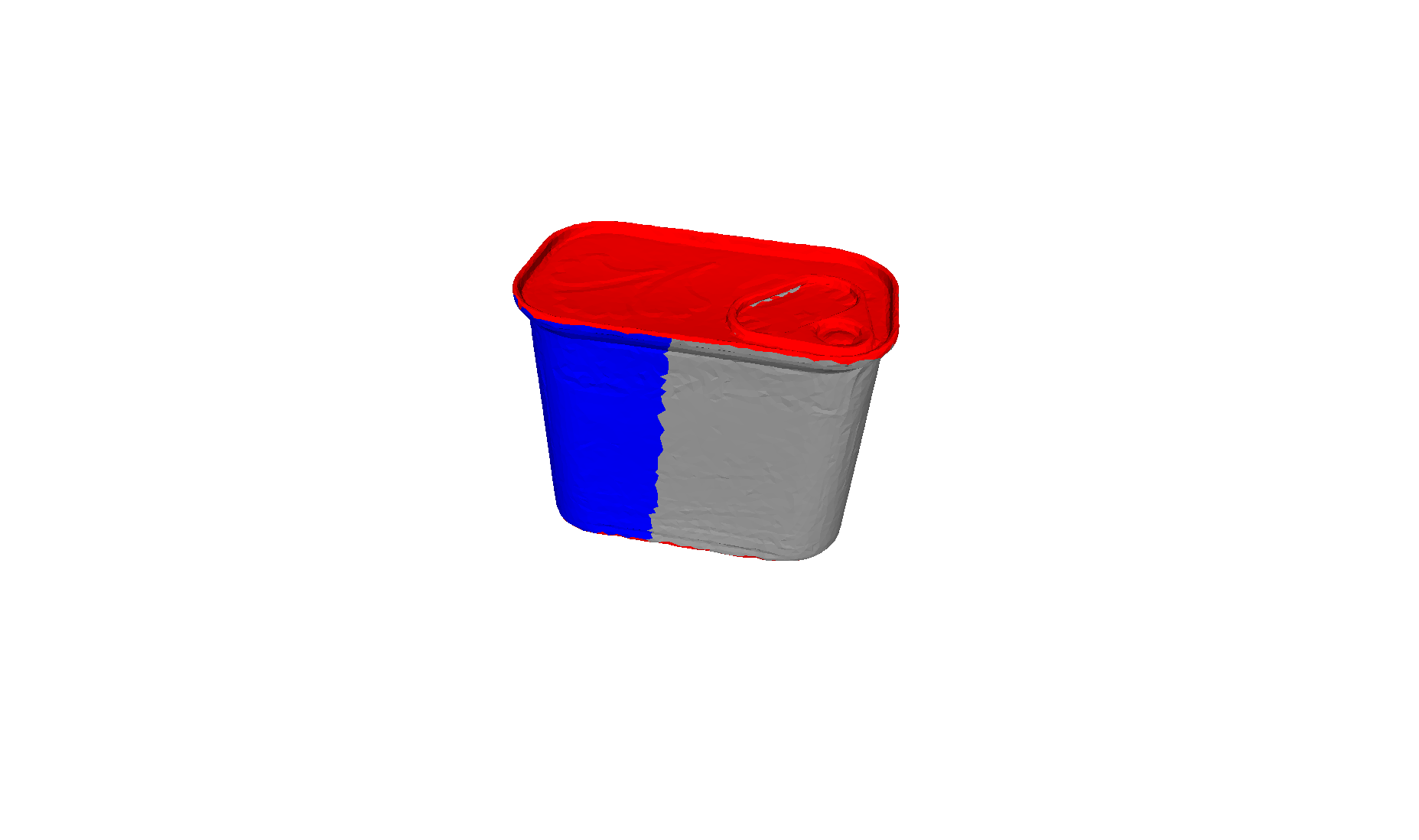}%
    }
    \subfloat{ %
        \includegraphics[trim={400 100 200 100},clip,width=0.12\textwidth]{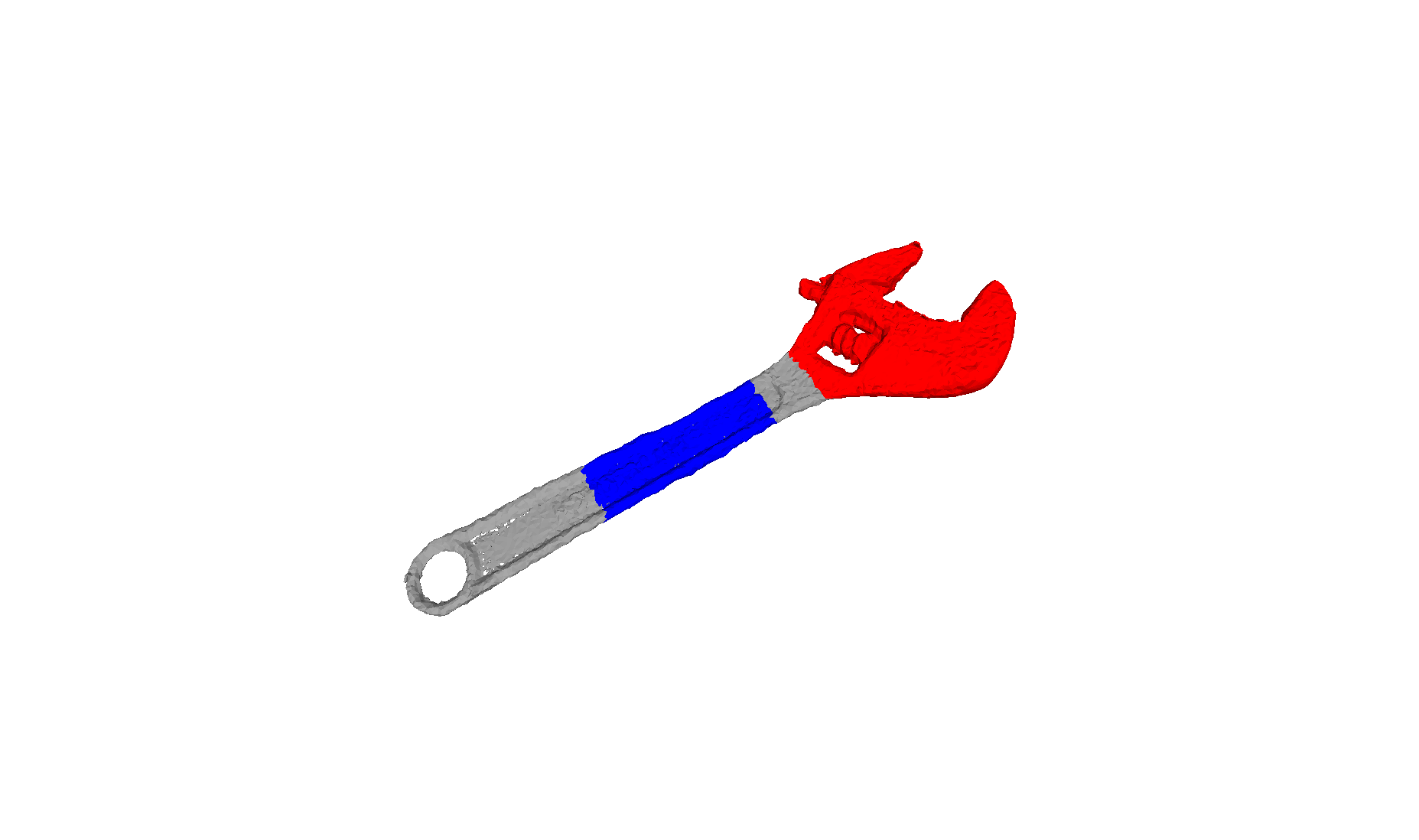}%
    }
\caption{Example contact regions defining in-hand manipulation tasks on YCB objects. The contact points on the object must be moved from the area in red to the area in blue.}
\label{fig:contact_regions}
\end{figure}

We describe the task with the aid of Fig.~\ref{fig:task_illustration}. An in-hand manipulation task is defined by:
\begin{itemize}
    \item An initial contact region, $C_i$, and a desired contact region, $C_d$, defined on the object's surface; examples of some of these contact regions on the YCB objects meshes are shown in Fig.~\ref{fig:contact_regions}.
    \item An initial grasp pose, $H_i$, and the desired grasp pose, $H_d$, identifying the hand's pose with respect to the object's reference frame.
\end{itemize}

The initial contact points, $P_i$, and the final contact points, $P_d$, between the hand and the object must lie inside the respective contact regions, in which they can be adjusted according to their feasibility with the hardware used.

Based on our literature review, we define three levels of performing an in-hand manipulation task, depending on which part of the desired configuration is targeted:
\begin{itemize}
    \item \textbf{Level I} desired hand pose~$H_d$ only;
    \item \textbf{Level II} desired contact region~$C_d$ only;
    \item \textbf{Level III} both~$C_d$ and~$H_d$.
\end{itemize}
With level~I, the hand can make contact at any region on the object and with level-II, the initial and final hand pose are not constrained. These different tasks allow for the exploration of different in-hand manipulation behaviors (e.g.\ finger gating vs in-hand rolling) without specifying how the manipulation should be performed. Additionally, we aim for our tasks to adapt to the needs and capabilities of different robots.

\subsection{Procedure}\label{sec:procedure}
For each task, the procedure runs as follows:
\begin{enumerate}
  \item The object is set at the initial grasp in the robot's hand~(e.g. a human places the object, the robot autonomously grasps the object, etc.). The contacts $P_i$ between the hand and the object must lie in the initial contact region $C_i$. The object pose at this initial pose is recorded as~$\hat{H}_i$. If the position error\%~(computed as $\frac{||s_i-\hat{s}_i||}{||sr_d-s_i||}\times 100$ or orientation error\% computed using Eq.~\ref{eq:or_perc}, between the setup initial hand pose~$\hat{H}_i$ and the initial hand pose~$H_i$ from the dataset is more than 10\%, this experiment must be discarded.
    \item The in-hand manipulation method is run, to move the object toward the desired configuration, defined by contact region~$C_d$ (or giving specific points $P_d$ in the region $C_d$), and by $H_d$. 
    \item The hand reaches a new pose on the object $H_r$ and new contacts $P_r$. This final configuration is recorded. The times for planning and executing are recorded.
\end{enumerate}

\subsection{Setup Description}
We defined several tasks for many of the YCB objects~\cite{calli_ycb}. These tasks contain initial and desired contact regions $C_i$, $C_d$, and are available in our protocol's associated website. The given contact regions are meshes that represent parts of the object's surface. If a method requires specific contact points to be defined instead of contact regions, researchers can define the corresponding contact points $P_i$, $P_d$ within these regions to match their hardware. In particular, the chosen YCB objects cover a wide spectrum in terms of size which enables benchmarking hands of different sizes.

The hand poses are strictly dependent on the robot's hand frame. We additionally provide some examples of $H_i$ and $H_d$ that identify the required change in grasp pose. The main objective is to obtain a relative change~$T$ between the initial grasp~$H_i^{rob}$ and the final grasp~$H_d^{rob}$ of your specific robot that matches the transformation~$T$ from $H_i$ to $H_d$~(i.e.$T = H_d H_i^{-1})$. Hence, the absolute hand poses for a specific robot should be adapted , but still impose the same relative change to the given frames ($H_d^{rob} = T H_i^{rob}$). 

Our examples work for end-effectors roughly the size of adult human hands, but most of them can be executed with bigger hands as well.

\subsection{Robot/Hardware Description}

This benchmark is targeted towards manipulation systems equipped with either a gripper or a multi-fingered hand. The tasks that we describe can also be performed with multiple arms, which could behave together as a dexterous hand. 

A perception system is required to estimate the initial and final pose of the object and the hand~(with pose of the links for level II and level III tasks). For computing the error metrics, we additionally require the mesh of the links of the hand. Given all this information, we provide software to estimate the errors as shown in Fig.~\ref{fig:geodesic}.

\subsection{Execution Constraints}
During the in-hand manipulation planning and execution (step 2 in section~\ref{sec:procedure}), a human cannot intervene once the object has been placed inside the robot's hand. The object should be in a stable grasp at the initial and final configurations. For the purpose of this benchmark, a grasp is stable if the hand pose with respect to the object is constant for 5 seconds after execution. During this time, the object must be in contact only with the robot's hand, without any help from external supports.

During the in-hand manipulation execution, the object can break contact with the robot. However, the contact cannot be broken to favour stable contacts with support surfaces. In fact, we do not allow several pick-and-place executions to change the grasp (i.e.\ regrasping~\cite{tournassound-icra1987}), because this does not classify as in-hand manipulation. In contrast, exploiting external surfaces to push the object inside the hand and performing non-prehensile manipulation is allowed, because the contact between the object and the robot is maintained while the object is in contact with the external surface. Similarly, the object can be thrown in the air and caught in a different configuration, since no contact with external surfaces happens once the robot releases the grasp. A second robot hand can be used as a support surface but not for holding the object, for regrasping between hands is not allowed.

%%% Local Variables:
%%% mode: latex
%%% TeX-master: "main"
%%% End:

\vspace{-0.5em}
\section{Benchmark Guidelines}\label{sec:guidelines}
\subsection{Scoring}
\label{sec:scoring}
We score the task success based on the error between the reached grasp~$G_r=[H_r,P_r]$ and the desired grasp~$G_d=[H_d,C_d]$. To ensure the robustness of a method, each grasp set needs to be run on the robot five times for the same object. We propose two separate sets of metrics to quantify error in hand pose and reached contact.

To quantify hand pose error, we split the hand pose~$H$ into position vector~$s$ and orientation quaternion $q$ and compute the error in position and orientation separately. Due to differences in the object and hand sizes, we also compute an error normalized by the distance between initial and final poses. The error between the reached hand position~$s_r$ and the desired hand position~$s_d$ is the Euclidean distance~$l_2$ norm
\begin{equation}
err_{pos}=||s_d-s_r||_2.
\end{equation}
To compute the percentage position error, we divide the error by the distance between the initial hand position~$s_i$ and desired hand position~$s_d$,
\begin{equation}
err_{pos}\%=100\times \frac{||s_d-s_r||_2}{||s_d-s_i||_2}. \label{eq:pos_perc}
\end{equation}
To measure the orientation error, we use the sum of quaternions~\cite{Huynh2009}. Given the desired hand orientation~$q_d$ and the orientation of the reached hand pose~$q_r$, the orientation error is defined as
\begin{equation}
err_{or}\%=100\times\frac{\min(||q_d-q_r||_2,||q_d+q_r||_2)}{\sqrt{2}}. \label{eq:or_perc}
\end{equation}
To quantify error in reaching the desired contact regions~$C_d$, we propose two metrics: Euclidean and Geodesic. We want all contacts made by the robot to be inside the desired contact region~$C_d$. Hence these metrics are used to find the largest distance~$max_d$ between the robot links and the desired contact region. We will first discuss these metrics to compute error between two points and then describe how to compute them between meshes.

Given two points, $p_1$ and $p_2$, the Euclidean metric~$G_{euc}(\cdot)$ is computed as the~$l_2$ norm
\begin{equation}
    G_{euc}=||p_2-p_1||_2.
\end{equation}
The Geodesic metric~$G_{geo}(\cdot)$ is the shortest path between the points on the object surface $\psi$
\begin{equation}
    G_{geo}=||p_2-p_1||_{\psi}.
\end{equation}
Efficient geodesic computation methods with software implementations are available~\cite{Crane_2017_HMD}.

For using these metrics with meshes, we propose the algorithm shown in Algorithm~\ref{al:contact}. We use the object mesh~$O$, the meshes of the~$N$~links of the robot~$L_{i \in N}$ making contact with the object, and the mesh of the desired contact region~$C_d$. To compute these metrics, we first project the desired contact region~$C_d$ and the robot meshes~$L_{i\in N}$ onto the object mesh~$O$ as shown in lines~\ref{alg:l:desired_vertices_start}-\ref{alg:l:hand_vertices_end}. Given the desired contact region~$C_d$, we compute the points~$P$ on the faces of~$C_d$ that are intersecting with the faces of the object mesh~$O$ using the function~\texttt{intersect($\cdot$)}. Many algorithms exist for performing this computation~\cite{gottschalk1996obbtree,larsen2000fast}. For every point in~$P$, we find a vertex in object mesh~$O$ with the \texttt{min-vertex($\cdot$)} function. This function finds the vertex that has the shortest Euclidean distance to the point. The set of vertices are stored as the desired contact vertices~$V_c$. This is also done for the robot meshes to obtain~$P_r$ (lines~\ref{alg:l:hand_vertices_start}-\ref{alg:l:hand_vertices_end}).

\begin{figure}
\centering
  \includegraphics[width=0.9\columnwidth]{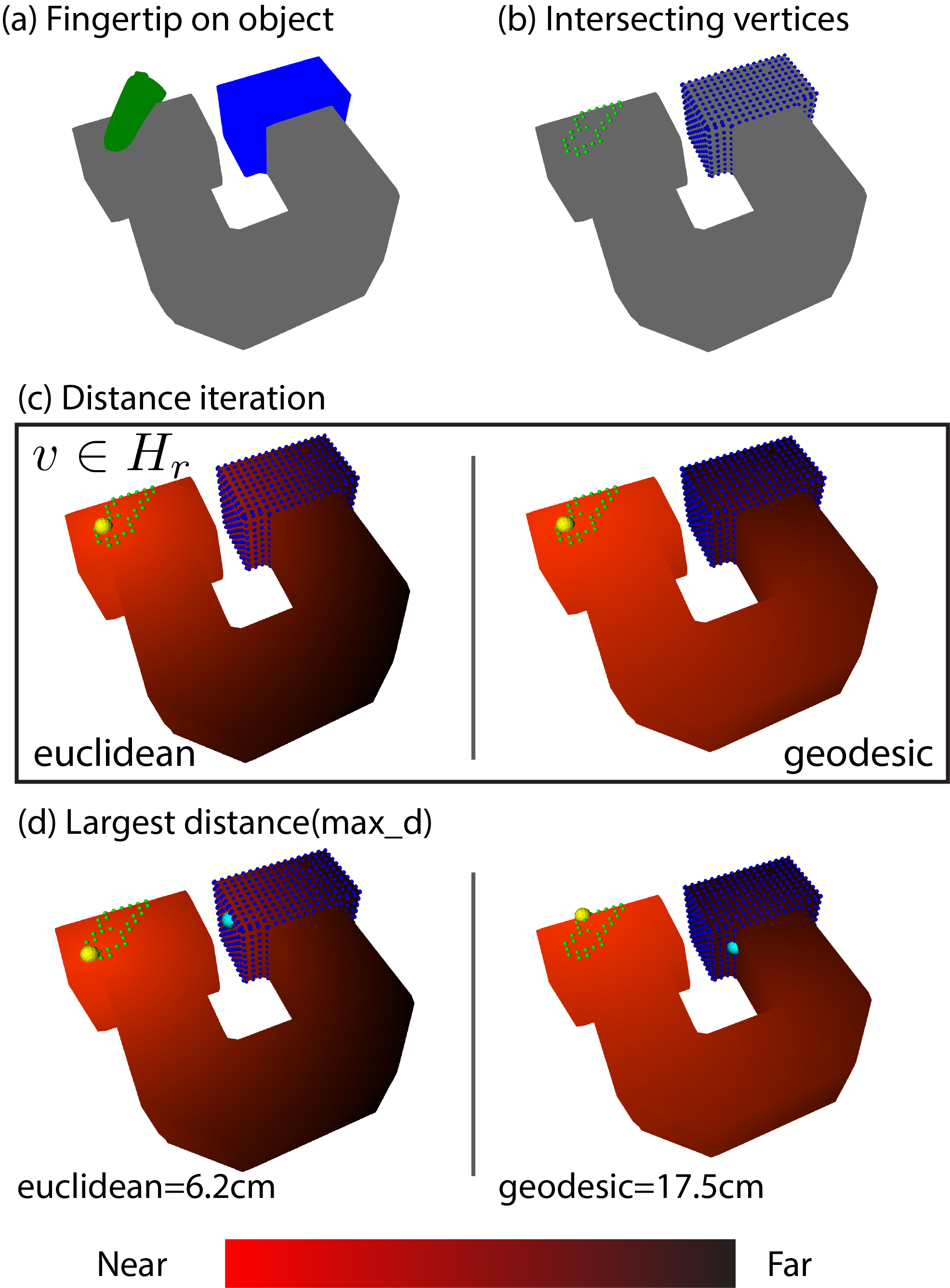}
  \caption{Computation of the contact region error from~Alg.~\ref{al:contact} is illustrated. The reached fingertip is shown as the green mesh on the object~(gray mesh) is shown in~(a) along with the desired contact region in blue. (b) shows intersecting vertices computation~(lines \ref{alg:l:hand_vertices_start}-\ref{alg:l:hand_vertices_end}).(c) shows the computation of distance from one vertex of the fingertip as a heatmap along the object surface. The shortest distance from the intersecting vertices of the fingertip~$P_r$ to the desired contact region~$V_c$ is shown in~(d). The cyan vertex in the desired contact region is the closest vertex to the robot hand.}
  \label{fig:geodesic}
\end{figure}

For each vertex in~$P_r$, we find the shortest distance to the desired contact set of vertices~$V_c$~(lines \ref{alg:l:max_contact_dist_start}-\ref{alg:l:max_contact_dist_end}). The distance function~$G(\cdot)$ in line~\ref{alg:l:distance_metric} is replaced by the Euclidean~$G_{euc}$ or geodesic~$G_{geo}$ metrics to report the Euclidean contact region error or Geodesic contact region error respectively. We find the vertex in~$P_r$ that is furthest from the desired contact region and report the distance as the error. These steps are visually illustrated in Fig.~\ref{fig:geodesic}. Since our contact metric is based on vertices on the object mesh, we also report the Euclidean distance between the two closest vertices in the object mesh as~$G_{min}$ to give readers an idea of accuracy of the contact error. We provide software to compute all of our proposed error metrics at our associated website.

\begin{algorithm}[t]
  \SetAlgoLined
  \DontPrintSemicolon
  \KwData{$O$,$L_{i\in N}$,$C_d$}
  \KwResult{max\_d}
  $P_r\gets$[]\;
  $V_c\gets$[]\;
  max\_d$\gets 0$\;
  $P \gets$ \texttt{intersect}($C_d$,$O$)\;
  \For{$v \in P$}{\label{alg:l:desired_vertices_start}
    $V_c$.append(\texttt{min-vertex}($v$,$O$))\;
    }\label{alg:l:desired_vertices_end}
  \For{$i \in N$ }{ \label{alg:l:hand_vertices_start}
    $V \gets$ \texttt{intersect}($L_i$,$O$)\;
    \For{$v \in V$}{  
       $P_r$.append(\texttt{min-vertex}($v$,$O$))\;
    }   
  }\label{alg:l:hand_vertices_end}
  \For{$v\in P_r$}{ \label{alg:l:max_contact_dist_start}
    min\_d$\gets\infty$\;
    \For{$p\in V_c$}{
      d$\gets G(v,p)$\;\label{alg:l:distance_metric}
      \If{\normalfont d$<$min\_d}{
        min\_d$\gets$d\;
      }
    }
    \If{ \normalfont min\_d$>$max\_d}{
      max\_d$\gets$min\_d\;
    }
  } \label{alg:l:max_contact_dist_end}
   \Return max\_d
   \caption{Contact region error computation~\label{al:contact}}
\end{algorithm}

\subsection{Details of Setup}
Researchers following the protocol need to report the robotic platform used and any additional information about the experimental setup. Object's weights, if different from what was given in~\cite{calli_ycb}, must be reported. Any noise in the perception system should be mentioned along with a discussion of if it affects the results, if any.

\subsection{Results to Report}\label{sec:results_to_report}
Researchers are encouraged to submit their initial and desired grasp sets along with object names for enabling others to use the same grasps for direct comparison. Additionally, we require the reporting of different values in the conducted experiments, which vary depending on the chosen task level to address:

\begin{enumerate}
    \item The error between the initial hand pose and the ``human-setup'' hand pose, computed as $err_{pos}$, $err_{or}$~\% (Level I, III), and $G_{euc}$, $G_{geo}$ (Level II, III).
    \item The error between the reached grasp and the desired grasp, computed as $err_{pos}$, $err_{pos}\%$, $err_{or}$\% (Level I, III), and $G_{euc}$, $G_{geo}$, as well as $G_{min}$ for accuracy of contact error (Level II, III).
    \item The time spent to plan and execute the in-hand manipulation method across the chosen object set. If the chosen method requires an offline planning step, we suggest to report it separately from the execution time.
    \item The percent of unsuccessful executions, specifying which trials failed and the cause (e.g. object dropped). 
\end{enumerate}

Apart from using these results to provide a quantitative analysis of their methods, we encourage researchers to submit the computed errors in the given website, where we provide code to visualize the errors as a box plot. In this box plot, the middle line defines the median error; the bottom and top borders indicate the first and third quartiles; the whiskers indicate the extremes of the inliers within 1.5 times the interquartile range. An example plot is shown in Fig.~\ref{fig:pose_error}.

\subsection{Comparing Results}

Given the measurements proposed in section~\ref{sec:results_to_report}, good methods have low errors, low planning and execution times, and low failure rates. Comparisons can be made between methods that address the same level of tasks, and level III can be compared with levels I and II with respect to their intersecting error types.
Systems with similar hardware (e.g. parallel gripper vs parallel gripper, 3-finger hand vs 3-finger hand) can rely on the same error comparison. Alternatively, The tradeoff between hardware complexity and result accuracy can be considered as an additional measure for comparing the results of different methods.

Due to the highly heterogeneous nature of the systems we aim at evaluating, it is difficult to find an \emph{absolute best}. As such, we believe users can compare the different methods according to the reported results and confront them with the requirements of the system. In particular, this allows for a choiche of the \emph{relative best} system for the particular application or requirements.

%%% Local Variables:
%%% mode: latex
%%% TeX-master: "main"
%%% End:

\section{Demonstration}
\label{sec:demo}
We benchmark methods for level I and level III tasks in Sec.~\ref{sec:level1} and Sec.~\ref{sec:level3} respectively. Since the level III task demonstrate the metrics used for level II tasks, we do not explicitly demonstrate a separate method for level II. We refer readers to~\cite{sundaralingam-icra2018-finger-gaiting} for an example approach that could be easily extended for evaluation using the level II protocol.
\vspace{-1.3em}
\subsection{Level I task}
\label{sec:level1}
Using the proposed framework, we compare five different in-hand manipulation solutions using Level~I tasks:
\begin{itemize}
\item The \emph{relaxed-rigidity, relaxed-position, \& relaxed-position-orientation} in-hand manipulation methods by Sundaralingam and Hermans~\cite{sundaralingam_in-grasp_manipulation}; these methods enable a robotic system to repose a grasped object without breaking/making new contacts on the object.
    \item The \emph{IK-rigid} method, in which a rigid contact model between the object and the fingertips is assumed.
    \item The \emph{point-contact} method, which assumes a point contact with friction model for the fingertips. That is, the contact position is assumed fixed, while the relative orientation can change. This is a  simplification of the model formulated in~\cite{Li1989}.
\end{itemize}
The desired grasp is given by a desired palm pose~$P_d$ with respect to the object.

\begin{figure}
	\centering
	\includegraphics[width=0.45\textwidth]{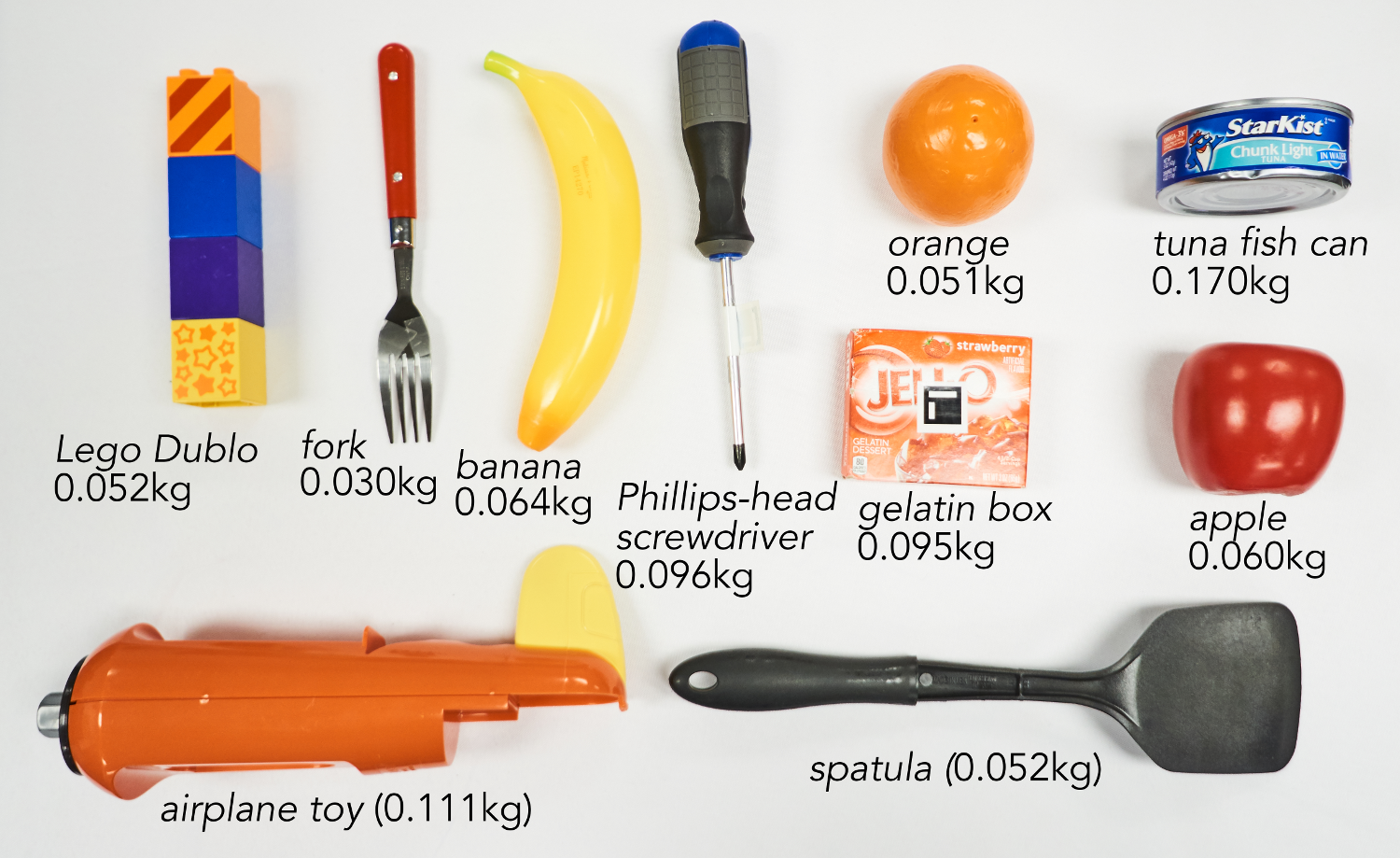}
	\caption{Objects from the YCB dataset used in the experiments with corresponding labels and measured weights below them.}
	\label{fig:obj}
\end{figure}

\subsubsection{Setup Details}
The methods are first compared within a trajectory optimization framework offline. Then, they are executed on the Allegro hand--a multi-fingered hand attached to a box frame. For the evaluation, we used ten objects from the YCB dataset shown in~Fig.~\ref{fig:obj}. The object and hand were tracked using ARUCO markers~\cite{Aruco2014} using an ASUS XTION camera. A human initialized the object in the initial hand pose. We setup a rigid transformation between the human setup pose and the planned initial pose to account for the human error in the reached pose. Each generated trajectory was executed five times. We used two different initial grasps and five different desired grasps per object. Five trials were run for each generated trajectory, accounting for 50 executions per object. In total, 500 trajectories were executed on the robot per method. The goal positions range from 0.8 to 8.33cm, with a mean of 4.87cm, from their respective initial positions. The goal orientations range from 1.53\% to 30.7\%, with a mean of 11.96\%, from their respective initial positions. The generated grasp sets are available at~\url{https://robot-learning.cs.utah.edu/project/in_hand_manipulation}.
% median position- 4.83, median orientation: 11.98
\iffalse
\begin{figure}
\centering
\includegraphics[width=0.48\textwidth]{rr_figs/p_time}
\caption{Comparison of time taken to generate the trajectory across five different methods: \emph{IK-rigid, point-contact, relaxed-position, relaxed-position-orientation} and \emph{relaxed-rigidity}.}
\label{fig:plan_time}
\end{figure}
\fi
\begin{figure}
\centering
\includegraphics[width=0.48\textwidth]{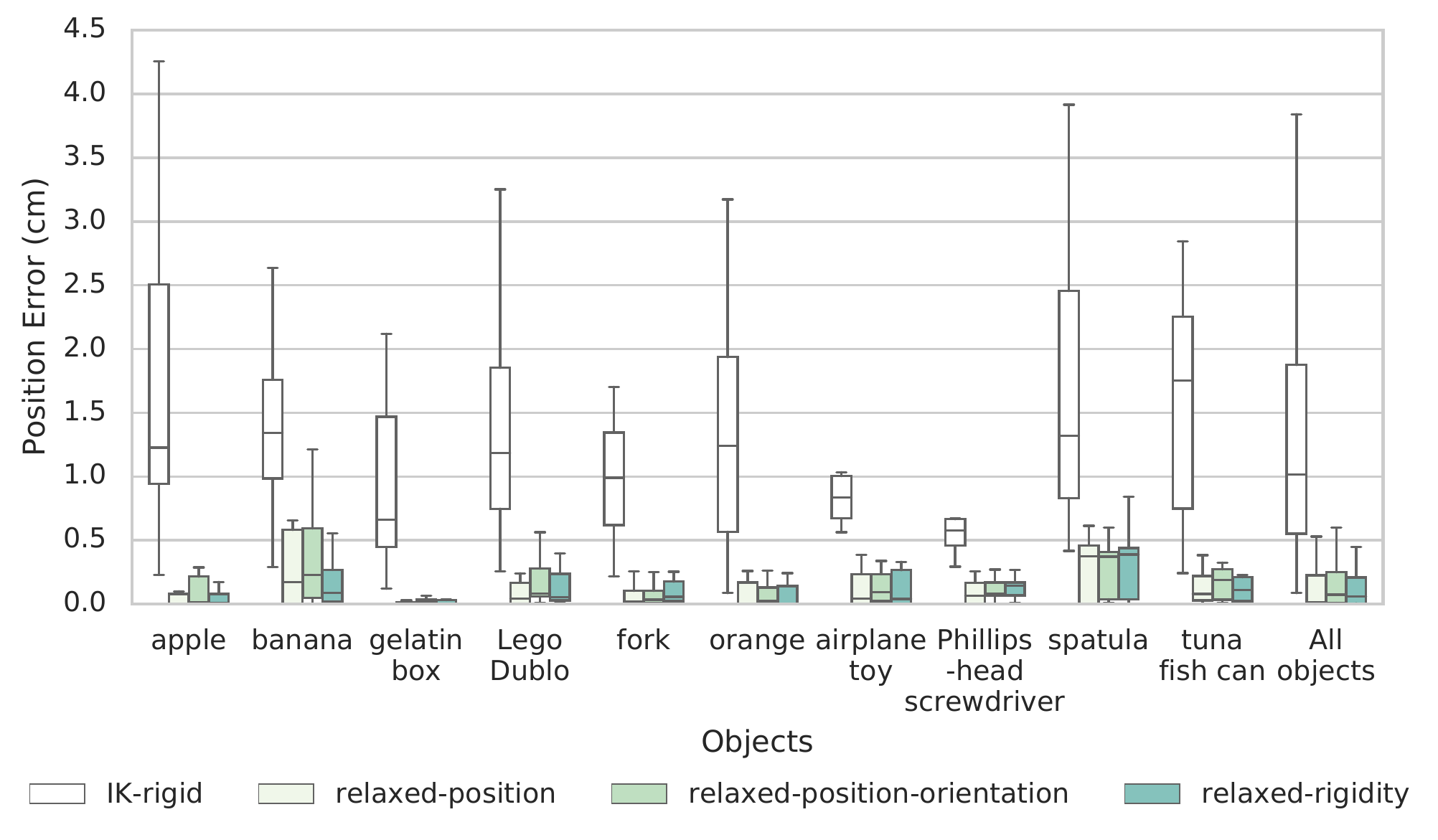}%
\caption{Comparison of planning results across four Level~I methods: \emph{IK-rigid, relaxed-position, relaxed-position-orientation} and \emph{relaxed-rigidity}. Results show the position error between the desired and expected final hand pose obtained by the planner~(error in planning).}
\label{fig:ds_error}
\end{figure}

\begin{figure}
  \centering
\subfloat{ %
\includegraphics[width=0.48\textwidth]{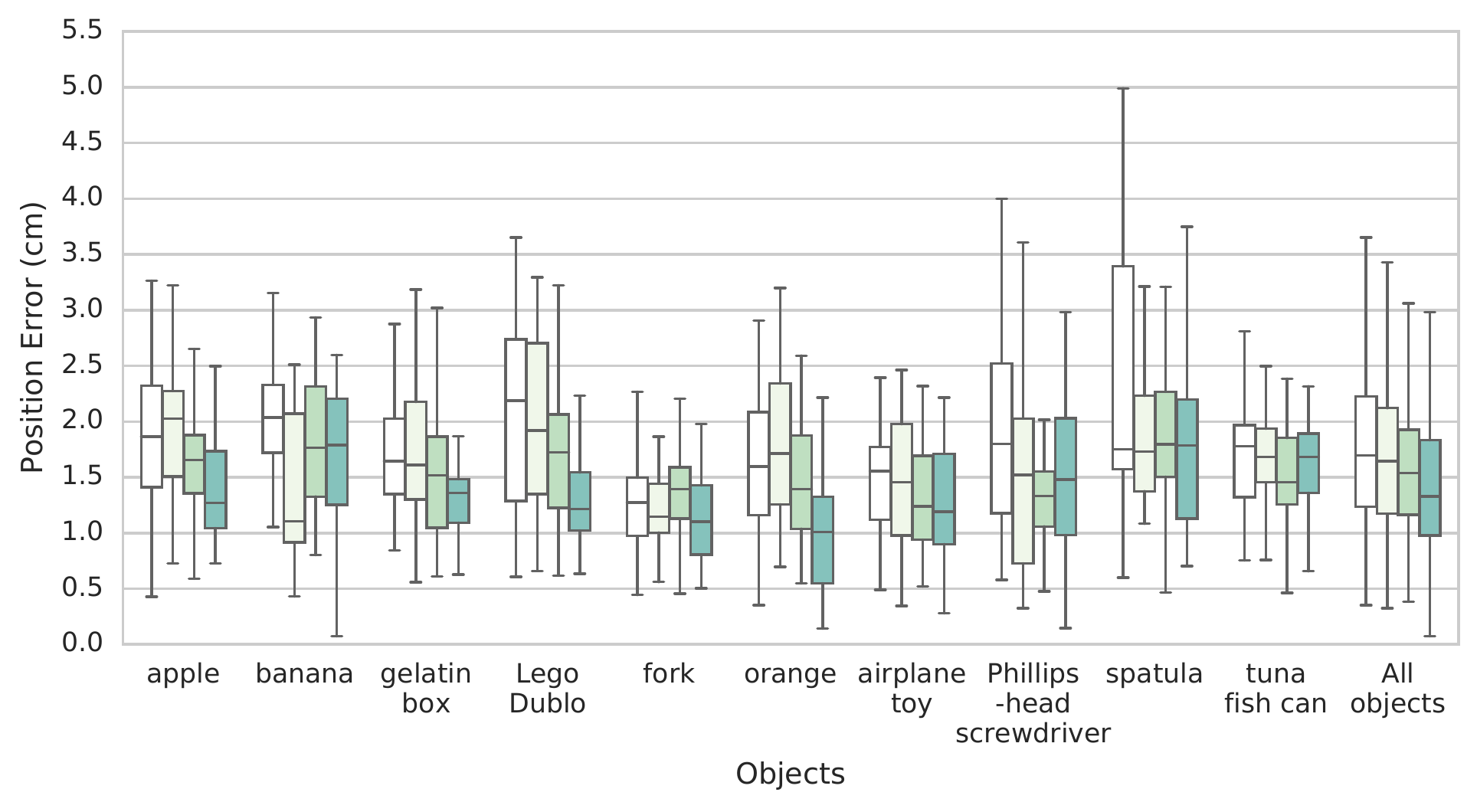} %
}\\[-0.01ex]
\subfloat{ %
\includegraphics[trim={0 0.8cm 0 0.0cm},clip,width=0.48\textwidth]{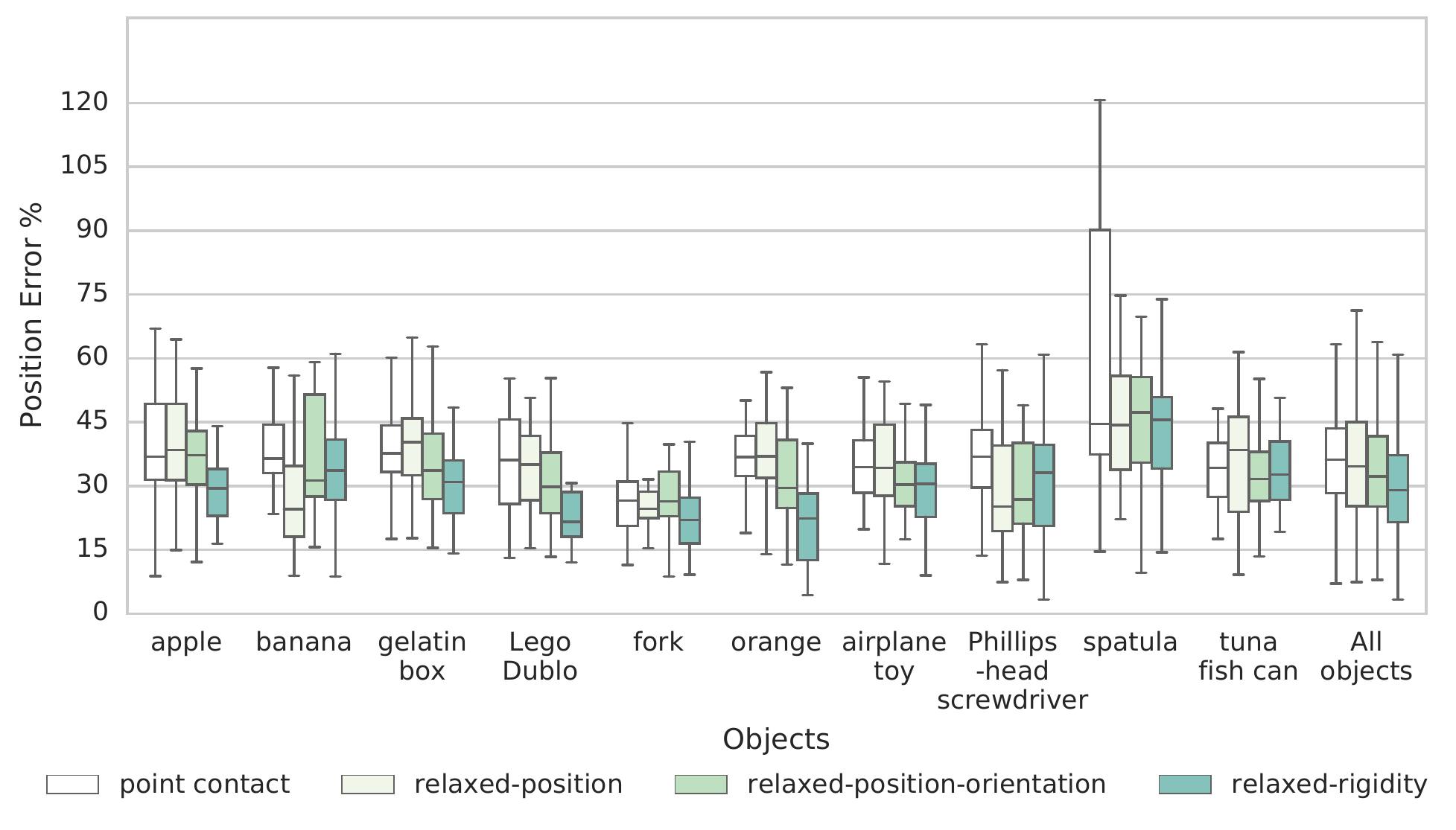} %
}\\[-0.01ex]
\subfloat{ %
  \includegraphics[trim={0 0 0 0.0},clip,width=0.48\textwidth]{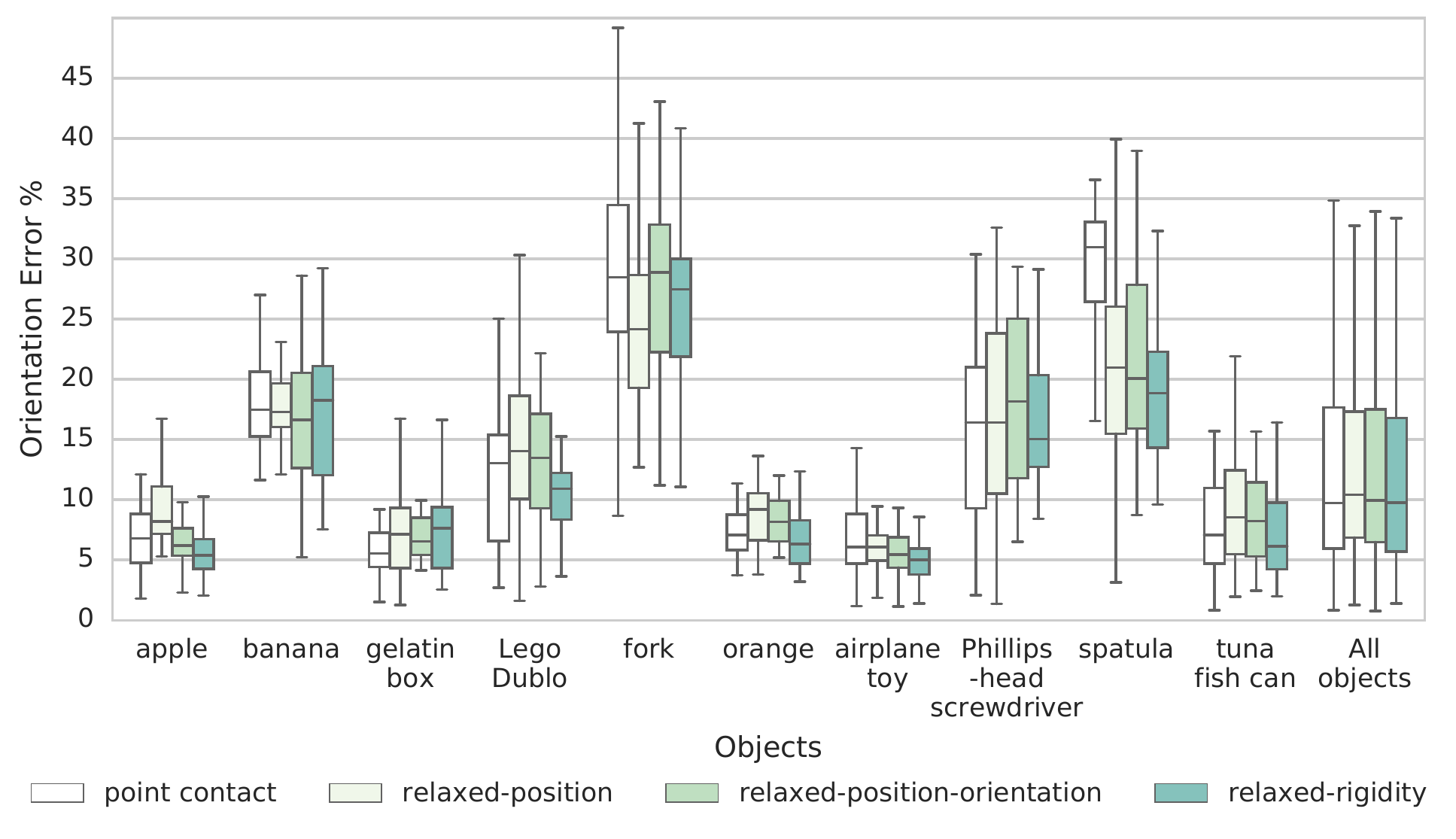}
}
\caption{A comparison of the different methods on real-world executions. Top: Position error Middle: Position error\% Bottom: Orientation error\%. The median position error decreases for all objects with the~\emph{relaxed-rigidity} method. Except for \emph{banana} and \emph{gelatin box}, the orientation error\% improves for the~\emph{relaxed-rigidity} method for all objects.}
\label{fig:pose_error}
\end{figure}

\subsubsection{Results}

For every trajectory that run on the robot, we record the position and orientation error. The median planning time across all objects were 14.8s, 4.4s, 0.5 s, 0.3s, and 0.5s for \emph{point-contact}, \emph{IK-rigid}, \emph{relaxed-position}, \emph{relaxed-position-orientation}, and \emph{relaxed-rigidity} respectively. Since all trajectories are run without replanning, the execution time is fixed at 1.67s. As suggested in our benchmarking framework, the errors are plotted as a box plot (showing first quartile, median error, third quartile) with whiskers indicating the extremes of the inliers within 1.5 times the interquartile range. In all plots results correspond to objects grasped with three fingers. We will first report the error between the planned hand pose and the desired hand pose, followed by results on executing the generated trajectories on the real robot.

We report the error between the planned hand pose and the desired hand pose across these methods: \emph{IK-rigid, relaxed-position, relaxed-position-orientation} and \emph{relaxed-rigidity}. We do not show offline results for the \emph{point-contact} method as computing the object pose from the solution is not possible since the optimization does not internally simulate the object's pose. However, we will report the results of~\emph{point-contact} method in the real-robot experiments. The errors are plotted in Fig.~\ref{fig:ds_error}. It is evident that IK-rigid has difficulty reaching the desired object position, a result of the problem being over-constrained, as such we do not report experimental results for this method on the real robot.

The position error and orientation error for all trials across all objects are shown in Fig.~\ref{fig:pose_error}. The \emph{relaxed-rigidity} method has the lowest median position error across all objects. Its maximum error across all objects is also much smaller than the \emph{point-contact} method. Additionally, one can see that the \emph{relaxed-rigidity} method has a lower variance in final position than the competing methods across nearly all objects. We report the median errors and the percentage of object drops in Table~\ref{tab:results}. The \emph{relaxed-rigidity} method never dropped any object across the 500 trials that were executed while all other methods dropped the object at least 5 times. The median errors~$[err_{pos},err_{or}\%]$ between initial pose and human setup initial pose are [0.58cm, 3.44\%] for \emph{point-contact}, [0.47cm, 3.19\%] for \emph{relaxed-position}, [0.52cm, 3.32\%] for \emph{relaxed-position-orientation}, and [0.54cm, 3.55\%] for \emph{relaxed-rigidity}. The contact points based metrics, $G_{euc}$ and $G_{geo}$, are not reported because these methods perform Level I tasks.
\begin{table}
  \centering
  \caption{Summary of results with the best value in bold text. The errors are the median of all trials.}
  \begin{tabular}{l  r r r r}
    \toprule
    \multirow{2}{*}{Method} & \multirow{2}{*}{drops\%} & \multicolumn{2}{c}{$err_{pos}$} & \multirow{2}{*}{$err_{or}\%$}\\ \cline{3-4}
    &   & (cm)     & \% &      \\ \toprule
    point-contact &  5  & 1.69 & 36.81 & \textbf{9.74}  \\ \midrule
    relaxed-position   & 9  & 1.64 & 30.95 & 10.43  \\ \midrule
    relaxed-position-orientation     & 7  & 1.54 & 29.19 & 9.84  \\ \midrule
    relaxed-rigidity   & \textbf{0} & \textbf{1.32} & \textbf{28.67} & 9.86  \\ \bottomrule
  \end{tabular}
  \label{tab:results}
\end{table}

%%% Local Variables:
%%% mode: latex
%%% TeX-master: "main"
%%% End:

\begin{table}[t]
  \centering
  \caption{Metrics for the DMG method.}
  \begin{tabular}{l r r r r}
    \toprule
    \multirow{2}{*}{Object} & \multirow{2}{*}{\emph{gelatin box}} & \multirow{2}{*}{\emph{cracker box}} & \multirow{2}{*}{\emph{spatula}} & \emph{potted meat} \\
    & & & & \emph{can}  \\
    \toprule
    $err_{pos}$ (cm) & 0.505 & 0.267 & 0.513 & 0.610\\ \midrule
    $err_{pos}\%$ & 9.9 & 2.7 & 8.8 & 11.0 \\ \midrule
    $err_{or}\%$ & 0.016 & 0.044 & 0.023 & 0.049 \\ \midrule
    $G_{euc}$ (cm) & 1.125 & 0.862 & 0.746 & 0.663 \\ \midrule
    $G_{geo}$ (cm) & 1.127 & 0.862 & 1.505 & 0.663 \\ \midrule
    $G_{min}$ (cm) & 0.013   &   0.057  & 0.034  &  0.004  \\ \midrule
    DMG time (s) & 10.312 & 15.467 & 13.406 & 18.295 \\ \midrule
    Plan time (s) & 0.023 & 0.004 & 6.7e-05 & 0.002 \\
    \bottomrule
  \end{tabular}
  \label{tab:DMG_results}
\end{table}
\vspace{-0.3em}
\subsection{Level III task}
\label{sec:level3}
In this section, we show example evaluations for tasks that involve both desired hand pose and desired contact points with the object. We use the Dexterous Manipulation Graph method (DMG), which is a planner for in-hand manipulation that is based on a graph representation of the object's surface, obtained through the object's discretization into small areas. The DMG contains information on possible motions of a finger on the object's surface. Through this graph, in-hand manipulation is planned as a sequence of rotations and translations. A detailed explanation of the method is found in~\cite{cruciani_dexterous_manipulation_graph}. Since the motion execution is treated separately from planning (e.g. it can use pushing against the environment, bi-manual pushing, etc.), we focus only on evaluating the planned solution.

\subsubsection{Setup Details}
The DMG planner is used to find an in-hand manipulation solution for an ABB Yumi smart gripper. We select some of the contact region tasks we defined in section~\ref{sec:protocol_design}, and defined $H_i$ and $H_d$ accordingly. Since the DMG defined in~\cite{cruciani_dexterous_manipulation_graph} is designed for parallel grippers, two contact points per hand are defined. We also define initial and desired poses $P_i$, $P_d$. Due to the gripper's structure, its position can be derived using a translation from the middle point between the two fingertips. All the executed tasks are available on our website.

\subsubsection{Results}
Table~\ref{tab:DMG_results} shows the results for planning in-hand manipulation paths. Each column corresponds to a different task with a different object. Each row shows the metrics proposed in section~\ref{sec:scoring}, and the planning time. The evaluated method requires offline computation of the Dexterous Manipulation Graph structure, reported as DMG time. Once the offline step is executed, the planning time for the given tasks is also reported.

%%% Local Variables:
%%% mode: latex
%%% TeX-master: "main"
%%% End:

\vspace{-0.3em}
\section{Conclusion}\label{sec:conclusion}
We proposed a benchmarking scheme for quantifying in-hand manipulation capabilities in a robotic system. We designed tasks for in-hand manipulation systems using the widely available YCB objects set, and we provided suggestions for adapting these tasks given the constraints of the hardware used for the evaluation. We have shown example results to demonstrate the outcome of the proposed benchmarking scheme. These results also serve as baselines for comparison with different methods in the future. By using this standardized evaluation we enable a comparison between different in-hand manipulation techniques that also considers different kinds of hardware platforms.
\vspace{-0.3em}
\bibliographystyle{IEEEtran}
\bibliography{In-Hand,Benchmarks}

%\addtolength{\textheight}{-12cm}   % This command serves to balance the column lengths
                                  % on the last page of the document manually. It shortens
                                  % the textheight of the last page by a suitable amount.
                                  % This command does not take effect until the next page
                                  % so it should come on the page before the last. Make
                                  % sure that you do not shorten the textheight too much.

%%%%%%%%%%%%%%%%%%%%%%%%%%%%%%%%%%%%%%%%%%%%%%%%%%%%%%%%%%%%%%%%%%%%%%%%%%%%%%%%

%%%%%%%%%%%%%%%%%%%%%%%%%%%%%%%%%%%%%%%%%%%%%%%%%%%%%%%%%%%%%%%%%%%%%%%%%%%%%%%%

%%%%%%%%%%%%%%%%%%%%%%%%%%%%%%%%%%%%%%%%%%%%%%%%%%%%%%%%%%%%%%%%%%%%%%%%%%%%%%%%

%%%%%%%%%%%%%%%%%%%%%%%%%%%%%%%%%%%%%%%%%%%%%%%%%%%%%%%%%%%%%%%%%%%%%%%%%%%%%%%%

\end{document}